\documentclass[runningheads]{llncs}

\usepackage{eccv}

\usepackage{eccvabbrv}

\usepackage{graphicx}
\usepackage{booktabs}

\usepackage[accsupp]{axessibility}  

\usepackage[pagebackref,breaklinks,colorlinks,citecolor=eccvblue]{hyperref}

\usepackage{orcidlink}

\usepackage{multirow}

\definecolor{Orange}{RGB}{237, 125, 49}

\begin{document}

\title{Diffree: Text-Guided Shape Free Object Inpainting with Diffusion Model} 
\titlerunning{Diffree}

\author{
Lirui Zhao\inst{1,2\dagger} \and
Tianshuo Yang\inst{2,3\dagger} \and
Wenqi Shao\inst{2\dagger\ddag} \and
Yuxin Zhang\inst{1} \and \\
Yu Qiao\inst{2} \and
Ping Luo\inst{2,3} \and 
Kaipeng Zhang\inst{2\ddag\star} \and
Rongrong Ji\inst{1\star}
}

\renewcommand{\thefootnote}{\fnsymbol{footnote}}
{\let\thefootnote\relax\footnotetext{
\noindent \hspace{-5mm}
$^\dagger$Equal contribution\; $\ddag$Project lead\; $^\star$Corresponding author
}}


\authorrunning{L.~Zhao et al.}

\institute{
$^1$Key Laboratory of Multimedia Trusted Perception and Efficient Computing, \\ \; Ministry of Education of China, Xiamen University \\
$^2$OpenGVLab, Shanghai AI Laboratory 
$^3$The University of Hong Kong \\
\url{https://github.com/OpenGVLab/Diffree}
}

\maketitle
\begin{figure}[!h]
  \vspace{-1.cm}
  \centering
  \includegraphics[width=0.9\linewidth]{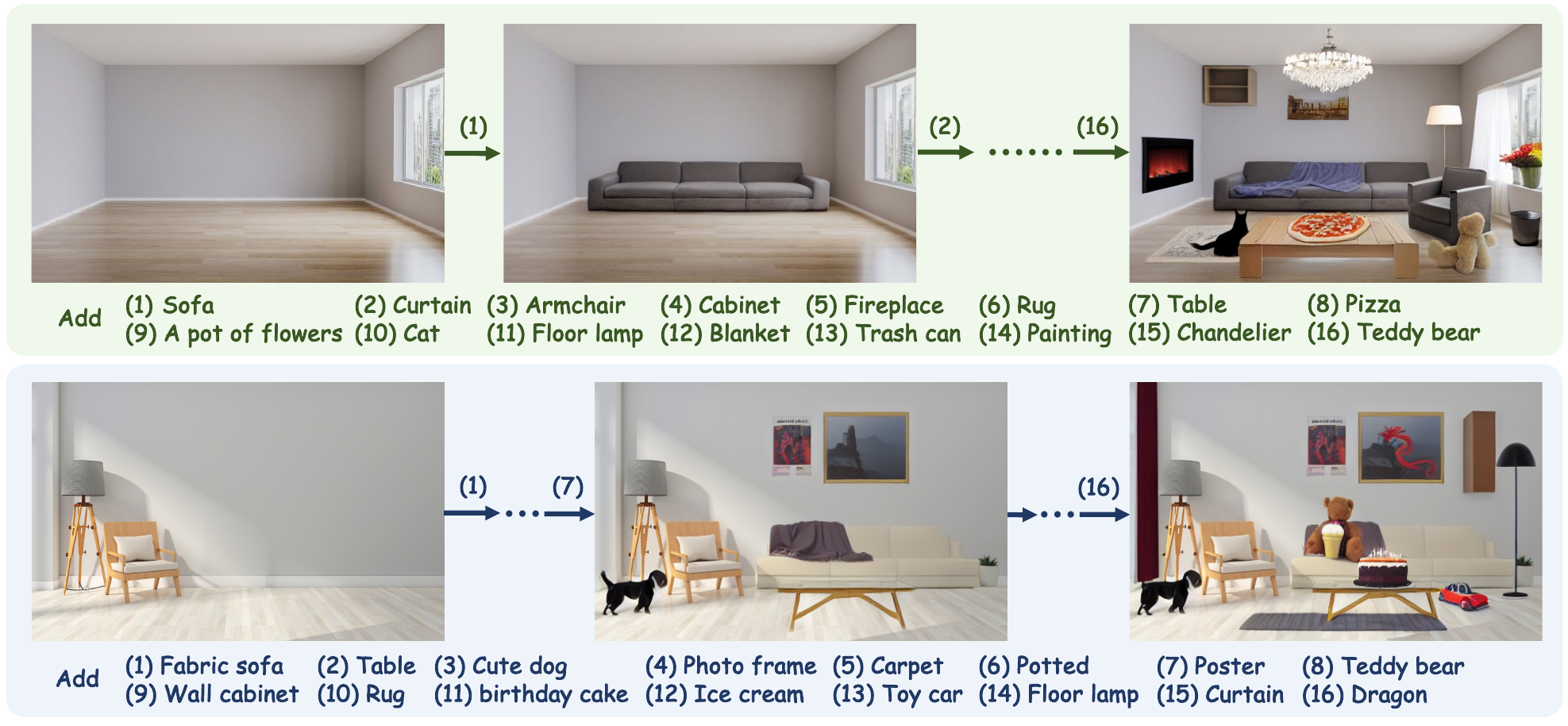}
  \vspace{-0.2cm}
  \caption{Our approach iteratively generates inpainting results. The objects from text guided is reasonably added in images while ensuring the background consistency.}
  \vspace{-1.2cm}
  \label{fig:iterative_addition_results}
\end{figure}
\begin{abstract}

This paper addresses an important problem of object addition for images with only text guidance. It is challenging because the new object must be integrated seamlessly into the image with consistent visual context, such as lighting, texture, and spatial location. While existing text-guided image inpainting methods can add objects, they either fail to preserve the background consistency or involve cumbersome human intervention in specifying bounding boxes or user-scribbled masks. To tackle this challenge, we introduce Diffree, a Text-to-Image (T2I) model that facilitates text-guided object addition with only text control. To this end, we curate OABench, an exquisite synthetic dataset by removing objects with advanced image inpainting techniques. OABench comprises 74K real-world tuples of an original image, an inpainted image with the object removed, an object mask, and object descriptions. Trained on OABench using the Stable Diffusion model with an additional mask prediction module, Diffree uniquely predicts the position of the new object and achieves object addition with guidance from only text. Extensive experiments demonstrate that Diffree excels in adding new objects with a high success rate while maintaining background consistency, spatial appropriateness, and object relevance and quality.


\end{abstract}
\section{Introduction}
\label{sec:intro}

With the recent remarkable success of Text-to-Image (T2I) models (\eg, Stable Diffusion~\cite{podell2023sdxl}, Midjourney~\cite{2022midjourney}, and DALL-E~\cite{betker2023dalle3,ramesh2022hierarchical}), creators can quickly generate high-quality images with text guidance. The rapid development has driven various text-guided image editing techniques~\cite{brooks2023instructpix2pix,geng2023instructdiffusion, zhang2024magicbrush,zhang2023hive,sheynin2023emu}. Among these techniques, text-guided object addition which inserts an object into the given image has attracted much attention due to its diverse applications, such as advertisement creation, visual try-on, and renovation visualization. While important, object addition is challenging because the object must be integrated seamlessly into the image with consistent visual context, such as lighting, texture, and spatial location.

\begin{figure}[!t]
  \vspace{-0.5cm}
  \centering
  \includegraphics[width=1.0\linewidth]{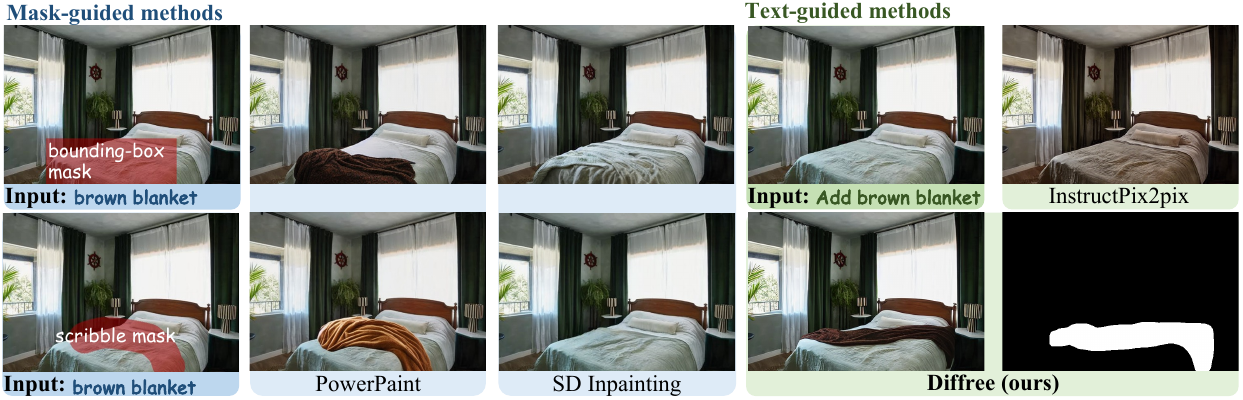}
  \vspace{-0.6cm}
  \caption{Qualitative comparisons of Diffree and different kinds of methods.}
  \vspace{-0.3cm}
  \label{fig:qualitative_results}
\end{figure}

Existing techniques for object addition in images can be broadly categorized into mask-guided and text-guided approaches (\cref{fig:qualitative_results}). Mask-guided algorithms typically require the specification of a region where the new object will be inserted. For example, traditional image inpainting methods~\cite{bertalmio2000image, suvorov2022resolution, lugmayr2022repaint, yu2018generative, pathak2016context} focus on seamlessly filling user-defined masks within an image to match the surrounding context. Recent advancements, such as PowerPaint~\cite{zhuang2023task}, have effectively incorporated objects into images given their shape and textual descriptions while maintaining background consistency. However, manually delineating an ideal region for all objects, considering shape, size, and position, can be labor-intensive and typically requires drawing skills or professional knowledge.
On the other hand, text-guided object addition methods, such as InstructPix2Pix~\cite{brooks2023instructpix2pix}, attempt to add new objects using only text-based instructions. Despite this, these methods have a low success rate and often result in background inconsistencies, as demonstrated in \cref{fig:qualitative_results} and \cref{fig:pix2pix_statistical_result}. Additionally, when employing InstructPix2Pix for iterative object addition, the quality of the inpainted image tends to degrade progressively with each step, as illustrated in \cref{fig:without_mix}.

To tackle the above challenges, we introduce Diffree, a diffusion model with an additional object mask predictor module that can predict an ideal mask for a candidate inpainting object and achieve shape-free object addition with only text guidance. 
Compared with previous works ~\cite{xie2023smartbrush,zhuang2023task,brooks2023instructpix2pix,geng2023instructdiffusion}, our Diffree has three appealing properties.
\textit{First}, Diffree can achieve impressive text-guided object addition results while keeping the background unchanged. In contrast, previous text-guided methods ~\cite{brooks2023instructpix2pix} struggle to guarantee this. 
\textit{Second}, Diffree does not require additional mask input, which is necessary for traditional mask-guided methods~\cite{xie2023smartbrush}. In real scenarios, high-quality masks are hard to obtain.
\textit{Third},
Diffree can generate the instance mask and thus can be further combined with various existing works~\cite{chen2023anydoor,openai2023gpt4v} to develop exciting applications. 
For example, Diffree can achieve image-prompted object addition when combined with AnyDoor~\cite {chen2023anydoor} and plan to add objects suggested by GPT4V~\cite{openai2023gpt4v}, as shown in Fig. \ref{fig:applications}.

Towards high-quality text-guided object addition, we curate a synthetic dataset named Object Addition Benchmark (OABench) which consists of 74K real-world tuples including an original image, an inpainted image, a mask image of the object, and an object description. 
The data curation process is illustrated in \cref{fig:data_framework}. Note that object addition can be deemed as the inverse process of object removal. We build OABench by removing objects in the image using advanced image inpainting algorithms such as PowerPaint ~\cite{zhuang2023task}. In this way, we can obtain an original image containing the object, an inpainted image with the object removed, the object mask, and the object descriptions. We use instance segmentation dataset COCO~\cite{gupta2019lvis,lin2014microsoft} as the source data, which has two benefits. First, the source image captures comprehensive natural scenes where the location and shape of one individual object often exhibit intrinsic alignment with the overall scene. It helps guarantee the reasonability of new objects' location. For instance, a monitor is commonly situated behind computer peripherals. Second, the ground-truth mask of the object already exists in the instance segmentation dataset, which can be directly utilized in removing objects with background consistency preserved. By contrast, InstructPix2Pix~\cite{brooks2023instructpix2pix} collects image pairs using proprietary T2I model~\cite{rombach2022high} under prompt pair with subtle modifications. While this approach maintains new objects' reasonability, it poses difficulties in preserving background consistency.

\begin{figure}[tb]
  \centering
  \includegraphics[width=1\linewidth]{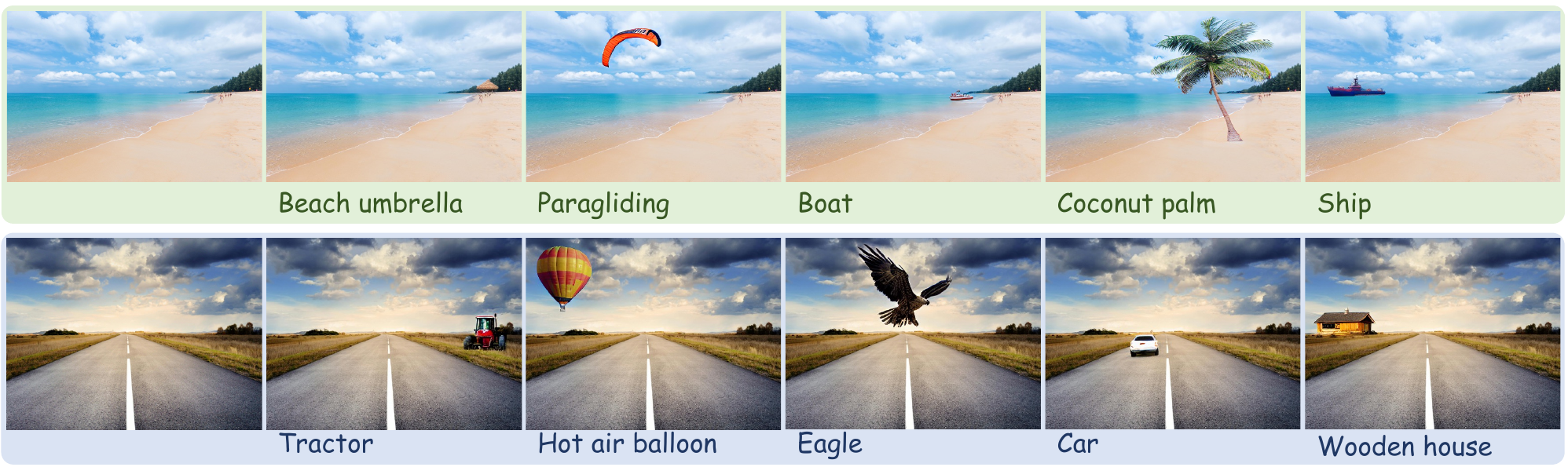}
  \vspace{-0.5cm}
  \caption{Diffree adds objects to the same image, with different spatial relationships.}
   \vspace{-0.4cm}
  \label{fig:different_in_one_image}
\end{figure}

With OABench, Diffree is trained to predict masks and images containing the new object given the original image and object text description. Thanks to the extensive coverage of objects in natural scenes in OABench, Diffree can add various objects to the same image while matching the visual context well as shown in \cref{fig:different_in_one_image}. Moreover, Diffree can iteratively insert objects into a single image while preserving the background consistency using the generated mask as shown in \cref{fig:iterative_addition_results} and \cref{fig:iterative_addition_process}.

For evaluation, we propose a set of evaluation rules through existing metrics~\cite{hessel2021clipscore,zhang2018unreasonable,heusel2017gans,xie2023smartbrush,openai2023gpt4v}, including consistency of background, reasonableness of object location, quality, diversity and correlation of generated object, and success rate. Extensive experiments show that Diffree performs better in object addition than previous mask-guided and text-guided techniques. For instance, Diffree obtains a significantly higher success rate than InstructPix2Pix. For successful cases, Diffree still outperforms InstructPix2Pix in various quantitative metrics.
%

The \textbf{contributions} of this work are three-fold.
1) We proposed Diffree, a model that can achieve text-guided shape-free object addition to free users from drawing the appropriate mask of objects.
The inpainted image from Diffree includes the new objects with reasonable shapes and consistent visual context.
2) We introduced OABench, an exquisite synthetic dataset for object addition. OABench comprises 74K real-world training data for the task of object addition. 
3) We evaluate this task with a set of rules for comprehensive assessment. Extensive experiments demonstrate the effectiveness of Diffree.
For example, Diffree achieves a high success rate (\eg, 98.5\% in COCO) and superior unified score (\eg, 38.92 versus 4.48) compared with other methods.

\section{Related Work}
\label{sec:relate}

\textbf{Text-to-Image Diffusion Models}
Recently, text-to-image (T2I) diffusion models~\cite{nichol2022glide,ramesh2022hierarchical,betker2023dalle3}, have shown exceptional capability in image generation quality and extraordinary proficiency in accurately following text prompts, under the dual support of large-scale text-image dataset~\cite{schuhmann2022laion} and model optimizations~\cite{dhariwal2021diffusion,ho2020denoising,rombach2022high}.
GLIDE~\cite{nichol2022glide} incorporated text conditions into the diffusion model and empirically showed that leveraging classifier guidance leads to visually appealing outcomes.
DALLE-2~\cite{ramesh2022hierarchical} enhances text-image alignment via CLIP~\cite{radford2021learning} joint feature space, DALLE-3~\cite{betker2023dalle3} further improves the prompt following abilities by training on highly descriptive generated image captions.
Stable Diffusion~\cite{rombach2022high}, which is well-established and widely adopted, garners significant attention and application within and beyond the research community.
Given that T2I models generate comprehensive images from text prompts, even minor alterations in prompts can result in substantial changes to the resultant image~\cite{brooks2023instructpix2pix}.
Consequently, there has been an increased focus not only on T2I generation but also on image editing based on additional conditions such as text inputs, masks, \etal.


{\setlength{\parindent}{0pt} 
\textbf{Text-Guided Image Editing}}
The effectiveness of the text-guided image editing methos~\cite{brooks2023instructpix2pix,zhang2024magicbrush,sheynin2023emu,geng2023instructdiffusion} largely depends on the composition of its dataset and how it is collected.
InstructPix2Pix~\cite{brooks2023instructpix2pix} combines two large pretrained models, a large language model~\cite{mann2020language} and a T2I model~\cite{rombach2022high}, to generate a dataset for training a diffusion model to follow written image editing text prompts.
Its innovative data collection method allows InstructPix2Pix to follow instructions and shows amazing effects, while makes its consistency is difficult to guarantee due to both input and output are generated by the T2I model.
InstructDiffusion~\cite{geng2023instructdiffusion} treats all computer vision tasks as image generation with multiple output formats, and aligns these tasks with human instructions. 
Emu Edit~\cite{sheynin2023emu} adapt its architecture for multi-task learning and train it an unprecedented range of tasks formulated as generative tasks, and displays strong and diverse results, especially in the combination of multiple tasks.
Emu Edit generate the output image and apply the mask-based attention to generate input iamge for object addtion task.
MagicBrush~\cite{zhang2024magicbrush} introduces a manually annotated dataset which both input and output are generated by the T2I model.
The image editing performance of fine-tuneing InstructPix2Pix on MagicBrush shows better.
Unlike the previous methods, we propose a novel and easily expandable collection method, thanks to the existing instance segmentation dataset, we use real images as output and synthetic images without a specific object as input.
Our work closely relates to the concurrent work PIPE~\cite{wasserman2024paint}, which independently explores similar concepts and methodologies. Both studies involve removing objects to collect an object addition dataset and train a diffusion model for text-guided object addition. Our approach additionally trains an Object Mask Predictor (OMP) module to predict the mask of objects. We believe that the concurrent exploration of these ideas underscores the significance and timeliness of this research direction.


{\setlength{\parindent}{0pt} 
\textbf{Mask-Guided Image Inpainting}}
Mask-guided image inpainting methods~\cite{xie2023smartbrush,wang2023imagen,zhuang2023task,chen2023anydoor} alter the image in specific areas under addition conditions (\eg, text), while maintaining the background in its original state.
SmartBrush~\cite{xie2023smartbrush} achieves precise object inpainting guided by text and mask through a novel training and sampling strategy .
Imagen Editor~\cite{wang2023imagen}, finetuned on Imagen~\cite{saharia2022photorealistic}, captures fine details in the input image by conditioning the cascaded pipeline to accomplish precise image inpainting through a user defined mask and text prompts.
AnyDoor~\cite{chen2023anydoor} employs a discriminative ID extractor and a frequency-aware detail extractor to characterize the target object, thereby facilitating effective object addition given an area and corresponding object image.
Powerpaint~\cite{zhuang2023task} demonstrates superior performance on various inpainting benchmarks attributed to the introduction of learnable tokens to distinguish different tasks.
Although these methods have achieved amazing image inpainting effects, their commonality is the need for a mask.
For ordinary users, drawing an object mask with appropriate shape, size, and aspect ratio, corresponding accurately to the object and image, presents an unignorable challenge.

\section{Methodology}
\label{sec:methodology}

Given an image and the object description, our goal is to add the object to the image while preserving the background consistency.
Following this, we initially introduce OABench, an synthetic dataset for this task, comprising image-text pairs (as input) with corresponding object masks and images containing the object (as output).
We provide an overview of our data collection pipeline in \cref{sec:oabench} and. 
We next present Diffree, an architecture amalgamating a Stable Diffusion model with an Object Mask Predictor (OMP) module in Sec. \ref{sec:diffree}.
The evaluation procedure is presented in \cref{sec:eval_metric}.

\begin{figure}[tb]
  \centering
  \includegraphics[width=1.0\linewidth]{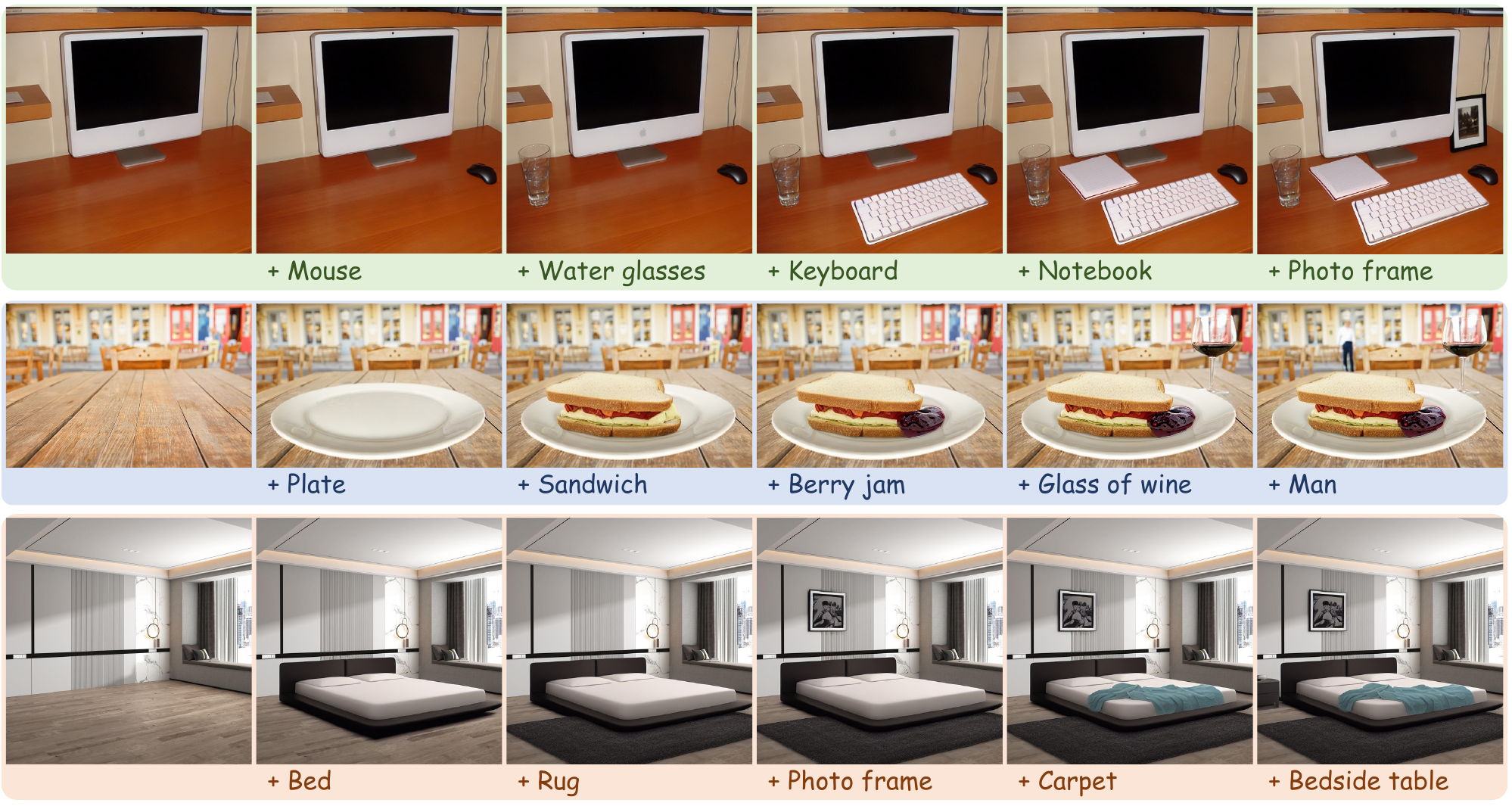}
  \vspace{-0.5cm}
  \caption{Diffree iteratively generates results. Objects added later can relate to the earlier.}
  \label{fig:iterative_addition_process}
\end{figure}

\subsection{OABench}
\label{sec:oabench}
We combine existing instance segmentation dataset~\cite{gupta2019lvis,lin2014microsoft} with powerful image inpainting method~\cite{zhuang2023task} to generate the OABench.
Unlike other instructions follow methods~\cite{brooks2023instructpix2pix,zhang2024magicbrush}, generating both data pairs using existing text-to-image (T2I) models~\cite{rombach2022high,ramesh2022hierarchical} with prompt pairs and filtering, we use the real image with object to synthesize the image without the object, as depicted in \cref{fig:data_framework}.
This can greatly ensure the consistency of the background as in other image inpainting methods~\cite{zhuang2023task} that require masks.
Furthermore, an object in the real image naturally aligns with its background, \ie, it is appropriate for generating the corresponding image without the same object.
In the following sections, we describe in detail the three steps of this process.

\subsubsection{Collection and Filtering}
\begin{figure}[tb]
  \centering
  \includegraphics[width=1.0\linewidth]{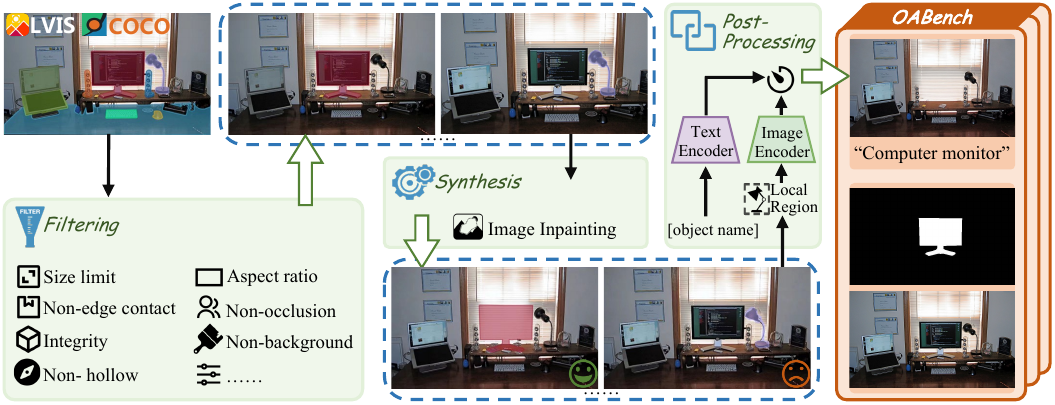}
  \vspace{-0.5cm}
  \caption{The data collection process of OABench.}
   \vspace{-0.6cm}
  \label{fig:data_framework}
\end{figure}

We gather and refine instances suitable for image inpainting by applying a set of rules from the LVIS dataset~\cite{gupta2019lvis}, a large instance segmentation dataset annotated for COCO~\cite{lin2014microsoft} dataset.
As depicted in \cref{fig:data_framework}, in images containing multiple instances, we enforce size constraints to exclude instances that are too big or too small (typically related to object components or background elements like buttons on clothing or rivers).
Subsequently, incomplete instances are filtered out using edge detection and integrity assessments.
Instances that are partially obscured are identified through cavity inspection, iterative IOU algorithm application, and common part comparison among various instances.
Additionally, objects with exceptionally high aspect ratios, which tend to yield subpar inpainting outcomes, are also eliminated.

\subsubsection{Data Synthesis}
\label{sec:synthesis}
We next employ a powerful image inpainting method, PowerPaint~\cite{zhuang2023task}, to eliminate specific instances obtained in the preceding stage.
%
%
Therefore, we can generate a synthetic image without specific objects with background consistency with the original image.
simultaneously, the original image, object mask, and corresponding name can be extracted from the LVIS and COCO.

\subsubsection{Post-Processing}
\label{sec:post_processing}
In the post-processing stage, we filter out the results with poor effects in image inpainting.
For some special cases (\eg, one of many dense and adjacent small cakes), image inpainting cannot effectively remove objects due to the complexity of the background.
Thus we calculate the clip score~\cite{hessel2021clipscore} using the object name and the region of the inpainted image, setting a threshold to remove images with higher scores which are deemed suboptimal.
Finally, OABench includes 74,774 high-quality data pairs, each data pair includes a synthetic image and object caption as input, object masks and original images as output.

\subsection{Diffree}
\label{sec:diffree}

\begin{figure}[tb]
  \centering
  \includegraphics[width=1.0\linewidth]{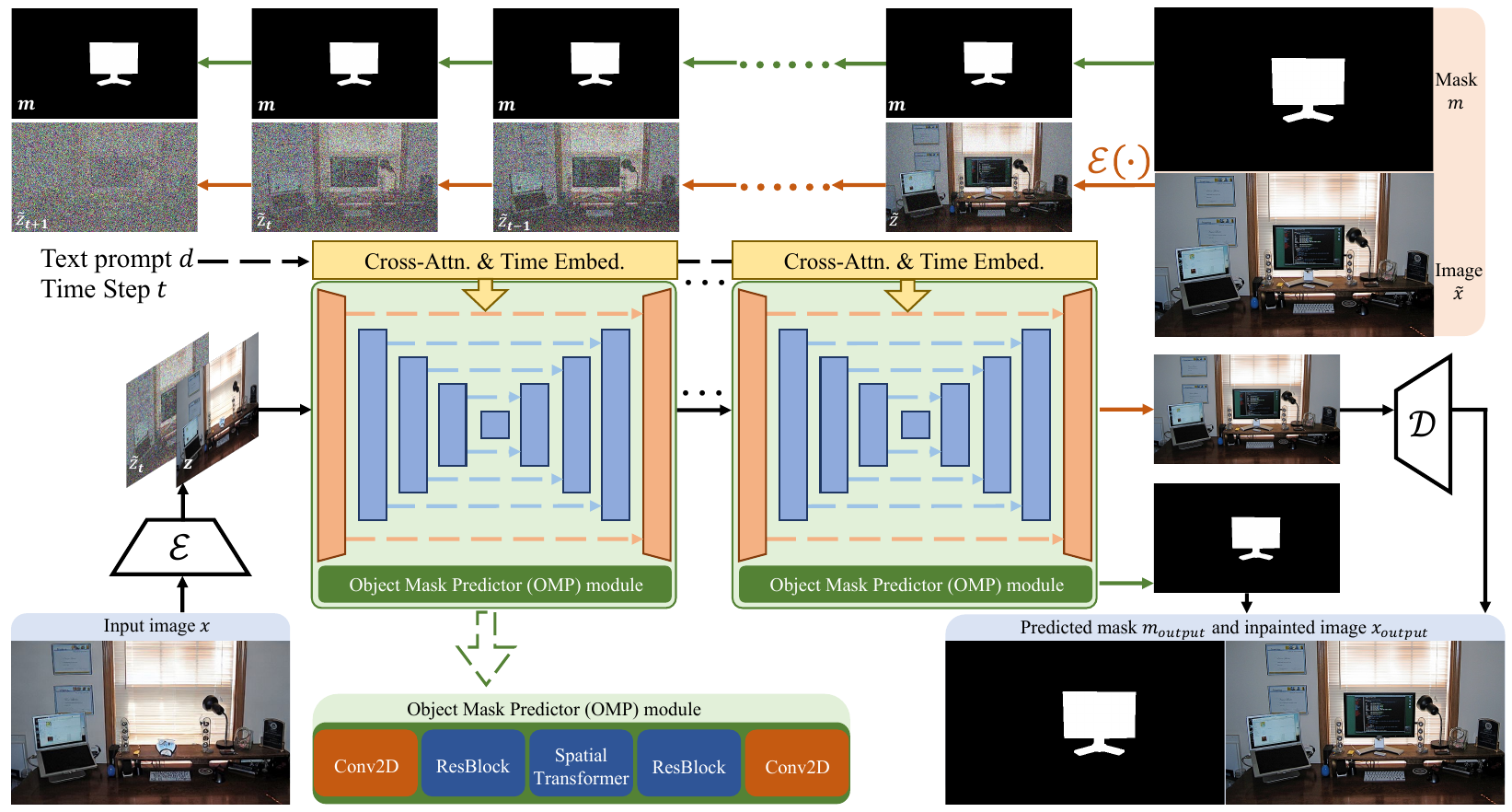}
  \vspace{-0.5cm}
  \caption{Overview of Diffree framework.}
   \vspace{-0.5cm}
  \label{fig:model_framework}
\end{figure}

For an image $x$ and a text prompt $d$, Diffree predicts a binary mask $m$ that specifies the region in $x$ and generates an image $\tilde{x}$. The masked region $\tilde{x} \odot m$ aligns with the text prompt $d$.
To this end, Diffree is instantiated with a pre-trained T2I diffusion model (\eg Stable Diffusion~\cite{rombach2022high}) with a object mask prediction (OMP) module as shown in \cref{fig:model_framework}.


\subsubsection{Diffusion Model} learns to generate data samples by iteratively applying denoising autoencoders that estimate the score function~\cite{song2020score} of a given data distribution~\cite{sohl2015deep}.
Stable Diffusion~\cite{rombach2022high} apply them in the latent space of powerful pre-trained variational autoencoder~\cite{kingma2013auto}, including encoder $\mathcal{E}$ and decoder $\mathcal{D}$, to reduce computing resources while maintaining quality and flexibility.
Stable Diffusion encompasses both forward and reverse processes. 
Given an image $\tilde{x}$, the forward process adds noise to the encoded latent $\tilde{z} = \mathcal{E}(\tilde{x})$:
\begin{align}
    \tilde{z_t}  = \sqrt{\bar{\alpha}_t}\tilde{z} + \sqrt{1-\bar{\alpha}_t}\epsilon, \epsilon \sim \mathcal{N}(0, \text{I})
    \label{eq:forward}
\end{align}
where $\tilde{z_t}$ is the noisy latent at timestep $t$, $\bar{\alpha}_t$ denotes the associated noise level.

In the reverse process, we learn a network $\epsilon_\theta$ that predicts the noise added to the noisy latent $\tilde{z_t}$, conditioned on both the image $x$ and text $d$.
To fine-tune Stable Diffusion for inpainting, we extend the channel of the first convolution layer to concatenate latent $z = \mathcal{E}(x)$ of image $x$ with $\tilde{z_t}$.
This allows Diffree to generate images by denoising step by step from Gaussian noise concatenated with the latent of the input image.
At the same time, the denoising process is guided by the associated feature $\mathrm{Enc}_{\rm txt}(d)$ of text $d$ encoded through the CLIP text encoder~\cite{radford2021learning}.
The network $\epsilon_\theta$ is optimized by minimizing the following objective function:
\begin{align}
L_{\rm DM} = \mathbb{E}_{\mathcal{E}(\tilde{x}), \mathcal{E}(x), d, \epsilon \sim \mathcal{N}(0, \text{I}), t }\Big[ \Vert \epsilon - \epsilon_\theta(\tilde{z_{t}}, z, \mathrm{Enc}_{\rm txt}(d), t) \Vert_{2}^{2}\Big].
\label{eq:loss_dm}
\end{align}

\subsubsection{OMP Module} 
\begin{figure}[tb]
  \centering
  \includegraphics[width=1.0\linewidth]{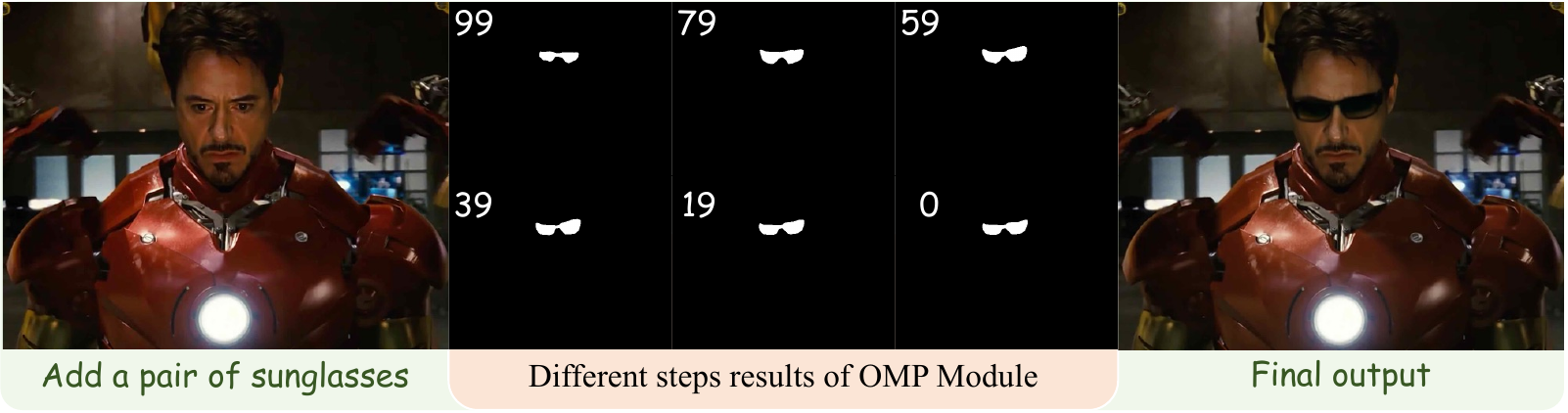}
  \vspace{-0.5cm}
  \caption{Visualization of masks from OMP at different steps of diffusion inference process (total 100 steps). The mask of added objects can be acquired in the very beginning.}
   \vspace{-0.2cm}
  \label{fig:mask_process}
\end{figure}

and diffusion model are trained simultaneously and used to predict the binary mask $m$.
The OMP module comprises two convolutional layers, two ResBlocks, and an attention block, as illustrated in \cref{fig:model_framework}.
First, we calculate the predicted noise-free latent $\tilde{o_t}$ using the output of the diffusion model:
\begin{align}
\tilde{o_t} = \frac{\tilde{z_{t}} - \sqrt{1-\bar{\alpha}_t}\epsilon_\theta(\tilde{z_{t}}, z, \mathrm{Enc}_{\rm txt}(d), t)}{\sqrt{\bar{\alpha}_t}}.
\label{eq:predicted_latent}
\end{align}
Here, the concatenation of $z = \mathcal{E}(x)$ with $\tilde{o_t}$ serves as inputs to the OMP module.
The gradient of $\tilde{o_t}$ is detached to optimize the two models without affecting each other.
We conduct bilinear interpolation downsampling on the mask $m$ to obtain $m'$, preserving its size identical to the input latent.
The OMP module’s network $\tau_\theta$ is optimized according to the following objective function:
\begin{align}
L_{\rm OMP} = \mathbb{E}_{\mathcal{E}(\tilde{x}), \mathcal{E}(x), d, m }\Big[ \Vert m' - \tau_\theta(\tilde{o_t}, z) \Vert_{2}^{2}\Big].
\label{eq:loss_omp}
\end{align}
It is noteworthy that the OMP module can predict the mask through the reverse process of diffusion rather than after it, as $\tilde{o_t}$ is available at each step, enabling mask prediction in the initial steps, as illustrated in \cref{fig:mask_process}.

We train both the diffusion model and the OMP module simultaneously. Combining \cref{eq:loss_dm,eq:loss_omp}, our ﬁnal training objective can be expressed as follows:
\begin{align}
L = L_{\rm DM} + \lambda L_{\rm OPS},
\label{eq:loss_total}
\end{align}
where $\lambda$ is a hyper-parameter which balances the two losses.

\subsubsection{Classifier-free Guidance}
Classifier-free diffusion guidance~\cite{ho2022classifier} is a method that involves the joint training of a conditional diffusion model and an unconditional diffusion model.
By combining the output score estimates from both models, this approach achieves a balance between sample quality and diversity.
Training for the unconditional diffusion model is achieved by fixing the conditioning value to a null variable intermittently throughout the training process.
We follow the approach of Brook~\etal~\cite{brooks2023instructpix2pix} by stochastically and independently defining our input conditions $x$ and $d$ as null variables with a probability of 5\%.

\subsection{Evaluation Metric}
\label{sec:eval_metric}

Due to the absence of robust quantitative metrics for shape-free object inpainting except the success rate, we propose a set of evaluation rules leveraging exits metrics~\cite{hessel2021clipscore,zhang2018unreasonable,heusel2017gans,xie2023smartbrush,openai2023gpt4v} to evaluate different methods in different aspects.

We first randomly select and manually inspect 1,000 evaluation data pairs from COCO~\cite{lin2014microsoft} and OpenImages~\cite{kuznetsova2020open} independently to ensure the validity of the object in the image and generalizability of the evaluation dataset. 
Each data pair comprises an original image $x_{\rm ori}$, a text prompt of an object $d$, and an inpainted image $x$.
The resulting output image $x_{\rm output}$ and the corresponding object mask $m_{\rm output}$ are outcomes derived from distinct methods.

\subsubsection{Background Consistency}
%
We adapt LPIPS~\cite{zhang2018unreasonable}, a widely adopted and robust metric for assessing the similarity between images, to evaluate this aspect:
\begin{align}
s_{\rm con}\left(x, x_{\rm output}, m_{\rm output}\right) = {\rm LPIPS}\left(x, x \odot m_{\rm output}  + x_{\rm output} \odot \left(1 - m_{\rm output}\right) \right).
\label{eq:score_consistency}
\end{align}

\begin{figure}[tb]
  \centering
  \includegraphics[width=1\linewidth]{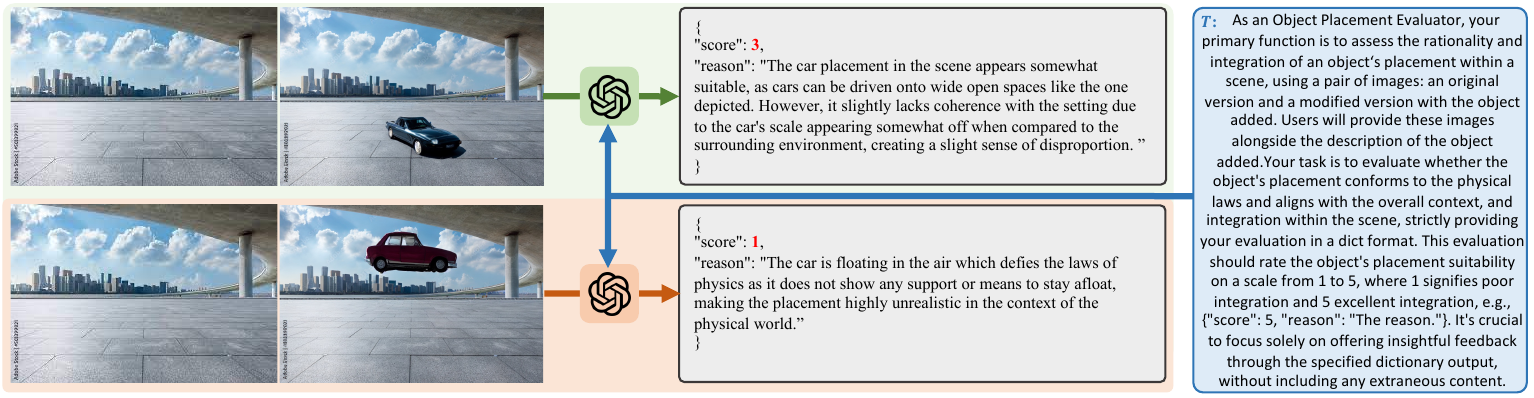}
  \vspace{-0.5cm}
  \caption{GPT4V shows good distinguish ability in the reasonableness between objects.}
   \vspace{-0.5cm}
  \label{fig:gpt_score}
\end{figure}

\subsubsection{Location Reasonableness}
Assessing the reasonableness of the object's location is a challenging task due to its inherent subjectivity.
Surprisingly, we note GPT4V~\cite{openai2023gpt4v} demonstrates strong discriminative abilities in assessing variations and evaluating different locations by providing $x$, $d$, $x_{\rm output}$ and an instruction $T$ as illustrated in \cref{fig:gpt_score}. GPT4V rates the appropriateness of the object's position on a scale from 1 to 5, while also providing justifications for these ratings:
\begin{align}
s_{\rm rea} \left(x, x_{\rm output}, d, T\right) = {\rm GPT4V}\left(x, x_{\rm output}, d, T \right)
\label{eq:score_reasonableness}
\end{align}

\subsubsection{Object Correlation }
To quantify this relationship, we utilize CLIP Score~\cite{hessel2021clipscore}, a metric to assess the correlation between text and image, by calculating the cosine similarity of their embeddings from CLIP~\cite{radford2021learning}.
we measure CLIP Score between the object area of $x_{\rm output}$ and $d$, which is referred to as “Local CLIP Score”:
\begin{align}
s_{\rm cor}(d, x_{\rm output}, m_{\rm output}) = {\rm CLIPScore}\left(d, \mathrm{Local}(x_{\rm output},m_{\rm output})  \right).
\label{eq:score_correlation}
\end{align}
where  $\mathrm{Local}(x,m)$ denotes obtaining a cropped region from $x$ using $m$.
To mitigate influences from background or mask shape, we compute an average of two Local CLIP Scores (one with background removal and another without).

\subsubsection{Object Quality and Diversity}
Following \cite{xie2023smartbrush}, we employ Local FID, measuring Fréchet Inception Distance (FID)~\cite{heusel2017gans} on the local regions, to evaluate the quality and diversity of generated object:
\begin{equation}
\begin{aligned}
s_{\rm qd}(LX_{\rm org}, LX_{\rm output}) = &{||\mu_{LX_{\rm org}} - \mu_{LX_{\rm output}}||}^2 +\\
&  \textrm{Tr}(\Sigma_{LX_{\rm org}} + \Sigma_{LX_{\rm output}} - 2*(\Sigma_{LX_{\rm org}}*\Sigma_{LX_{\rm output}})^{\frac{1}{2}})
\label{eq:score_quality_diversity}
\end{aligned}
\end{equation}
where $LX_{\rm org}$ and $LX_{\rm output}$ respectively denote the sets comprising all local regions of the original images and output images, $\mu$ and $\Sigma$ represent the mean and variance of the feature vectors obtained through a particular network~\cite{heusel2017gans}.

\subsubsection{Unified Metric}
Drawing upon the evaluation metrics delineated above (\cref{eq:score_consistency,eq:score_correlation,eq:score_reasonableness,eq:score_quality_diversity}), we compute a unified score to holistically assess text-guided shape-free object inpainting. 
We treat the derivative of inverse metric results (LPIPS and Local FID) as positive metrics and normalized the outcomes across different methods for each metric.
Ultimately, we average these normalized scores and multiply them by the success rate as a unified score.
The Unified metric not only considers success rate but also focuses on quantitative performances.

\section{Experiment}
\label{sec:experiment}

We comprehensively evaluated our model, Diffree, by conducting experiments on two benchmark datasets: COCO~\cite{lin2014microsoft}, and OpenImages~\cite{kuznetsova2020open}. Given the distinct input-output characteristics of our method compared to previous approaches, a quantitative comparison proves challenging. We align previous methods by adding auxiliary conditions, as depicted in \cref{sec:experimental_baselines}, and provide quantitative comparison results (\cref{sec:main_results}) to prove the effectiveness of Diffree more intuitively.
%
%
We then showcase visualizations of generated images and give corresponding analyses to offer an intuitive assessment of Diffree’s capabilities and comparisons in \cref{sec:visualization}.
finally, we demonstrate some applications to prove that Diffree is highly compatible with existing methods (\cref{sec:applications}).

\begin{figure}[tb]
  \centering
  \includegraphics[width=0.8\linewidth]{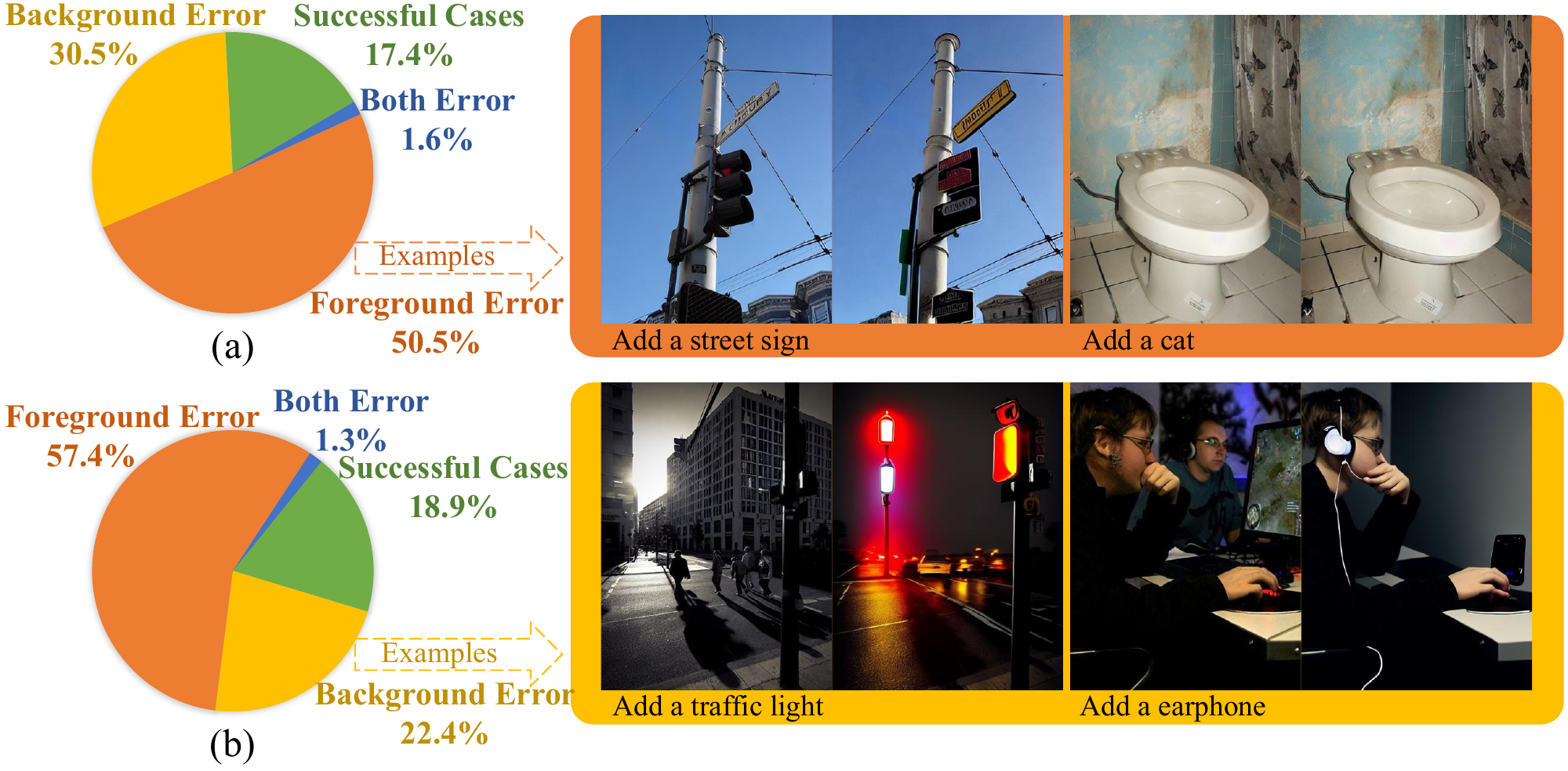}
   \vspace{-0.2cm}
  \caption{InstructionPix2Pix's result statistics in object addition. (a): COCO, (b): OpenImages. Foreground Error denotes failing to add objects or transforming existing objects, Background Error denotes inconsistent background.}
   \vspace{-0.5cm}
  \label{fig:pix2pix_statistical_result}
\end{figure}

\subsection{Experimental Settings}
\label{sec:experimental_settings}
\subsubsection{Training Setups}
we employ OABench to train Diffree, initializing the diffusion model with the Stable Diffusion 1.5~\cite{rombach2022high} weights.
We set $\lambda = 2$ in \cref{eq:loss_total} and set a batch size of 256.
Our model was trained around 10K steps on 8 A100 GPUs.

\subsubsection{Evaluation Datasets and Metrics}
As outlined in \cref{sec:eval_metric}, we employ four metrics (LPIPS~\cite{zhang2018unreasonable}, GPT4V~\cite{openai2023gpt4v}, Local CLIP Score and Local FID~\cite{xie2023smartbrush}) alongside the unified metric for evaluation on COCO~\cite{lin2014microsoft} and OpenImages~\cite{kuznetsova2020open}.


\subsubsection{Baselines}
\label{sec:experimental_baselines}
%
To facilitate comparison with prior methods~\cite{brooks2023instructpix2pix,zhuang2023task}, we manually check and annotate the object masks for InstructPix2Pix~\cite{brooks2023instructpix2pix}, and utilize our generated mask for PowerPaint~\cite{zhuang2023task} for generation. 
It is important to note that neither of these methods can complete evaluations independently.
thus, their quantitative metrics should be used as references only.


\begin{table}[tb]
\centering
\small
\caption{Main results on COCO and OpenImages. *: only calculate the successful cases' results. \dag: use the masks from our Diffree as PowerPaint's input.}
\vspace{-0.3cm}
\label{tab:main_results}
\begin{tabular}{llccc}
           \toprule 
           &                  & InstructPix2pix~\cite{brooks2023instructpix2pix} & PowerPaint~\cite{zhuang2023task}    & Diffree 
           \\ \midrule
\multirow{7}{*}{COCO~\cite{lin2014microsoft}}           
           & Success rate     & 17.4             & N/A           & \textbf{98.5}  \\ \cmidrule{2-5} 
           & LPIPS $\downarrow$          & \color{gray}{0.11*}          & \textbf{0.06} & 0.07           \\
           & GPT4V Score $\uparrow$            & \color{gray}{3.13*}          & N/A           & \textbf{3.47}  \\
           & Local CLIP Score $\uparrow$ & \color{gray}{29.30*}         & 28.74         & \textbf{28.96} \\
           & Local FID $\downarrow$      & \color{gray}{156.25*}        & 58.08         & \textbf{57.43} \\ \cmidrule{2-5} 
           & Unified Metric $\uparrow$   & 4.48            & \color{gray}{37.20\dag}      & \textbf{35.92}\\ \midrule\midrule
\multirow{7}{*}{OpenImages~\cite{kuznetsova2020open}}  
           & Success rate     & 18.9             & N/A           & \textbf{98.0}  \\ \cmidrule{2-5} 
           & LPIPS $\downarrow$            & \color{gray}{0.11*}            & \textbf{0.06} & 0.07           \\
           & GPT4V Score$\uparrow$            & \color{gray}{3.36*}            & N/A           & \textbf{3.50}  \\
           & Local CLIP Score $\uparrow$ & \color{gray}{29.21*}           & 28.57         & \textbf{28.81} \\
           & Local FID $\downarrow$        & \color{gray}{143.82*}          & 62.40         & \textbf{60.07} \\ \cmidrule{2-5} 
           & Unified Metric $\uparrow$   & 5.04            & \color{gray}{36.41\dag}      & \textbf{35.47} \\ \bottomrule
\end{tabular}
\vspace{-0.5cm}
\end{table}
\subsection{Main results}
\label{sec:main_results}
\cref{tab:main_results} shows the main results of Diffree with different baselines. We report the results of four powerful metrics and Unified Metric. 
It is worth to highlight that \textit{only successful InstructPix2pix results are computed and PowerPaint is utilized for image inpainting under the masks provided by the results of our approach}.
We can deduce the following conclusions from the results in several aspects.

\subsubsection{Success Rate}
We achieved a success rate of over 98\% on different dataset, while InstructPix2pix shows a lower success rate in object addition (17.2\% and 18.9\%).
As shown in \cref{fig:pix2pix_statistical_result}, most of the results of InstructPix2pix involve replacing existing object, without adding or significant changes to the background.
This demonstrates our excellent ability to complete this task.
Meanwhile it is not applicable to PowerPaint as it necessitates a mask input.

\subsubsection{Consistency of Background}
Diffree significantly outperforms InstructPix2pix in the LPIPS scores across all datasets (all decreased by 36\% than InstructPix2pix).
In particular, only scores from carefully chosen successful cases of InstructPix2pix were computed, potentially leading to an overestimation.
Furthermore, Diffree, as a shape free inpainting method, yields LPIPS results comparable to PowerPaint, as a shape required inpainting method. 
As discussed in \cref{sec:oabench}, we expect achieving consistency of background like the image inpainting methods that necessitate masks.
These methods inherently excel on this aspect, given that their input and ground truth are the same image during the training process.
Therefore, we believe that we have a strong capability in this aspect.

\subsubsection{Reasonableness of object location}
The results of GPT4V's assessment demonstrate that Diffree has a considerable advantage in the reasonableness of object location (\eg, 0.34 higher than successful results from InstructPix2pix).
This is not avaliable for PowerPaint due to it requires a mask as input.

\subsubsection{Correlation, Quality and Diversity of Generated Object}
We conduct an evaluation of the generated object across these three dimensions, utilizing both Local CLIP Score and Local FID.
Although Diffree exhibits a slightly lower Local CLIP Score in comparison to InstructPix2pix (\eg, 28.96 versus 29.30 on the COCO), this discrepancy can be rationalized by the fact that its successful results are inherently highly correlated while ours encompass all outcomes without any specific selection.
Intriguingly, we demonstrate superiority over PowerPaint in terms of correlation. 
Furthermore, our performance according to the Local FID metric indicates a distinct advantage relative to all other methods.

\subsubsection{Unified Metric of Diffree}
We combine the success rate with diverse metrics across various aspects to calculate a unified metric, thereby facilitating a more comprehensive comparison with extant text-guided methods.
It is discernible that Diffree exhibits a substantial superiority over InstructPix2pix, for instance, ours' 35.92 as opposed to InstructPix2pix's 4.48 on the COCO.
PowerPaint achieves superior results (\eg, 37.20 on the COCO dataset), the image inpainting of powerpaint buit upon the masks from our Diffree.
This further underscores the excellent scalability of Diffree when integrated with other methods.

\begin{figure}[tb]
  \centering
  \includegraphics[width=1\linewidth]{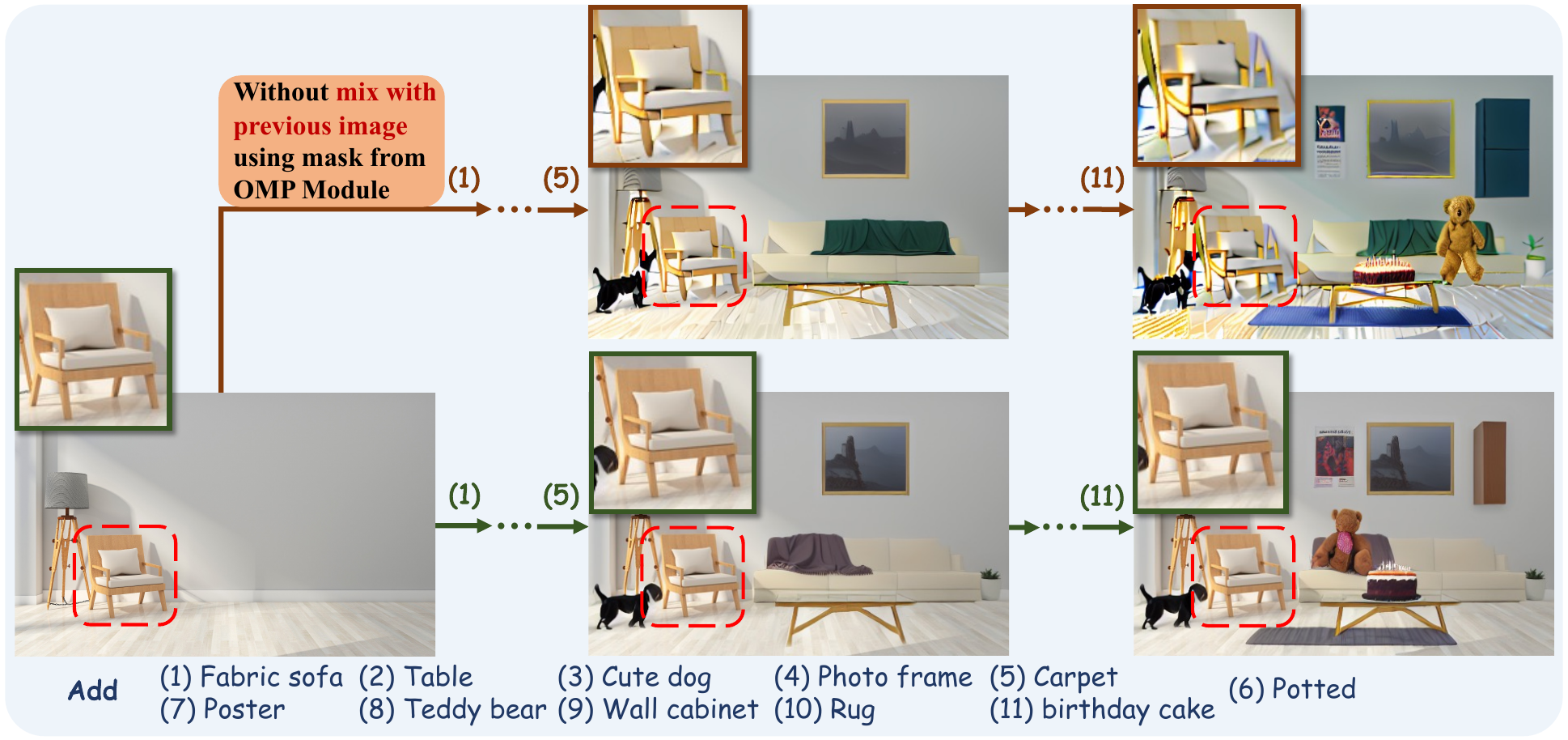}
  \vspace{-0.5cm}
  \caption{Visualization of with/without mix with previous image using mask from OMP.}
   \vspace{-0.5cm}
  \label{fig:without_mix}
\end{figure}


\subsection{Visualization}
\label{sec:visualization}
We provide different types' visualizations to more intuitively evaluate Diffree's capabilities~\cref{fig:iterative_addition_results,fig:iterative_addition_process,fig:different_in_one_image,fig:mask_process,fig:without_mix,fig:applications,fig:qualitative_results}, please refer to the respective image captions for detailed explanations. \textbf{For more results, please refer to the appendix.}

\subsection{Application}
\label{sec:applications}
Diffree can be well combined with other methods for more expansion.

\subsubsection{With GPT4V}
GPT4V~\cite{openai2023gpt4v} has a good ability to perceive and understand images, therefore we can use GPT4V for planning a object suitable for the image scene, seeing \cref{fig:applications}.
However, when task with adding corresponding object without altering the background, DALL-E-3~\cite{betker2023dalle3} in GPT4, falls short.

\subsubsection{With Other Methods}
\begin{figure}[tb]
  \centering
  \includegraphics[width=1\linewidth]{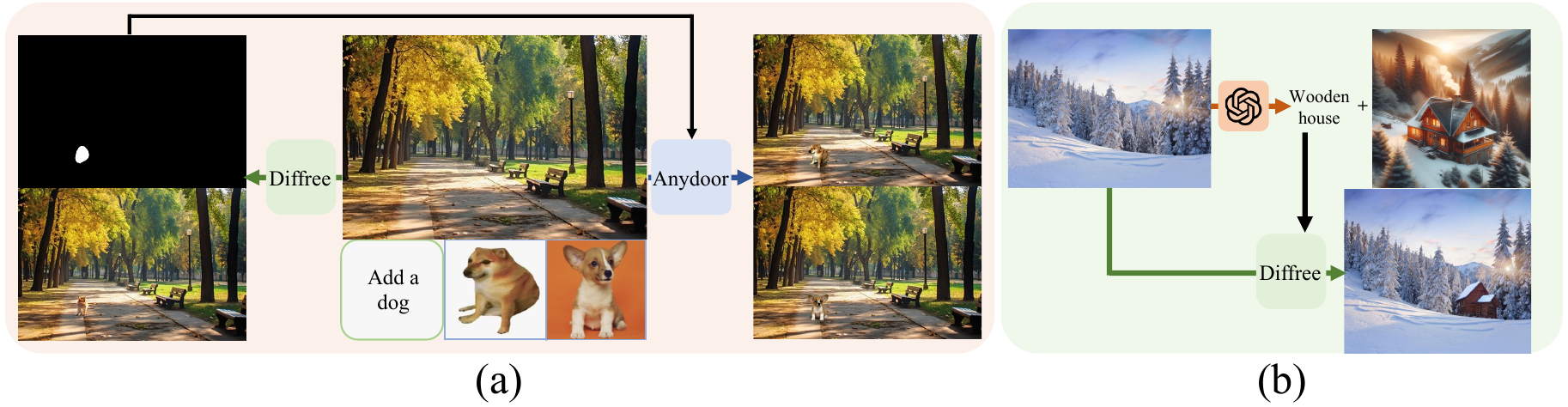}
  \vspace{-0.5cm}
  \caption{Applications combined with Diffree. (a): combined with anydoor to add a specific object. (b): using GPT4V to plan what should be added.}
   \vspace{-0.5cm}
  \label{fig:applications}
\end{figure}

AnyDoor~\cite{chen2023anydoor} can add a specific object to the designated area by providing a mask and object image. As depicted in \cref{fig:applications}, Diffree can combined with AnyDoor to further achieve adding a specific object to image.
DIffree also can effectively leverage the continuous progress in the image inpainting, to generate superior images, as demonstrated in \cref{tab:main_results}.

\subsubsection{Iterative Operation}
In \cref{fig:without_mix,fig:iterative_addition_results}, we present results of iterative inpainting. 
Leveraging the predicted mask from OMP module, Diffree can preserve the image background from cumulative degradation during successive inpainting. This holds potential applications within architectural and interior design domains.

\section{Conclusion}
\label{sec:conclusion}

We propose a novel method, Diffree, that leverages a diffusion model with an object mask predictor for text-guided object addition.
Beyond the method, we build a high-quality synthetic dataset, OABench, through a novel data collection method for this task.
Diffree distinguishes itself by preserving background consistency without requiring additional masks, which solves shortcomings of previous text-guided and mask-guided object addition methods. 
The quantitative and qualitative results demonstrate the superiority of our method.


%
%
\bibliographystyle{splncs04}
\bibliography{main}
\end{document}